\documentclass[conference]{IEEEtran}

\usepackage{algorithm,algorithmicx}
\usepackage{algpseudocode,empheq,hhline}

\usepackage{amsmath,amssymb,url,graphicx,array,color}
\usepackage{todonotes}
\usepackage{multirow}
\usepackage{balance}
\usepackage{fancyhdr}
\usepackage{url}
\urldef{\mailsa}\path{emails}

\DeclareMathOperator*{\argmax}{argmax}

\def\w{\omega}

\begin{document}

\title{Evolving Game Skill-Depth using General Video Game AI Agents}


%
%
%


\author{
\IEEEauthorblockN{Jialin Liu}
\IEEEauthorblockA{University of Essex\\
Colchester, UK\\
\url{jialin.liu@essex.ac.uk}}
\and
\IEEEauthorblockN{Julian Togelius}
\IEEEauthorblockA{New York University\\
New York City, US\\
\url{julian.togelius@nyu.edu}}
\and
\IEEEauthorblockN{Diego P\'erez-Li\'ebana}
\IEEEauthorblockA{University of Essex\\
Colchester, UK\\
\url{dperez@essex.ac.uk}}
\and
\IEEEauthorblockN{Simon M. Lucas}
\IEEEauthorblockA{University of Essex\\
Colchester, UK\\
\url{sml@essex.ac.uk}}
}

\maketitle

          \thispagestyle{plain}
          \fancypagestyle{plain}{
            \fancyhf{} 
            \fancyfoot[L]{978-1-5090-4601-0/17/\$31.00~\copyright2017~IEEE} 
            \renewcommand{\headrulewidth}{0pt}
            \renewcommand{\footrulewidth}{0pt}
          }
\begin{abstract}
Most games have, or can be generalised to have, a number of parameters that may be varied in order to provide instances of games that lead to very different player experiences.  The space of possible parameter settings can be seen as a search space, and we can therefore use a Random Mutation Hill Climbing algorithm or other search methods to find the parameter settings that induce the best games.  One of the hardest parts of this approach is defining a suitable fitness function.  In this paper we explore the possibility of using one of a growing set of General Video Game AI agents to perform automatic play-testing.  This enables a very general approach to game evaluation based on estimating the skill-depth of a game. 
Agent-based play-testing is computationally expensive, so we compare two simple but efficient optimisation algorithms: the Random Mutation Hill-Climber and the Multi-Armed Bandit Random Mutation Hill-Climber. For the test game we use a space-battle game in order to provide a suitable balance between simulation speed and potential skill-depth. Results show that both algorithms are able to rapidly evolve game versions with significant skill-depth, but that choosing a suitable resampling number is essential in order to combat the effects of noise.

\end{abstract}


\section{Introduction}


Designing games is an interesting and challenging discipline traditionally demanding creativity and insight into the types of experience which will cause players to enjoy the game or at least play it and replay it. There have been various attempts to automate or part-automate the game generation process, as this is an interesting challenge for AI and computational creativity~\cite{togelius2008experiment,browne2010evolutionary,cook2012aesthetic}.   So far the quality of the generated games (with some exceptions) do not challenge the skill of human game designers. This is because the generation of complete games is a more challenging task than the more constrained task of generating game content such as levels or maps.
Many video games require content to be produced for them, and recent years have seen a surge in AI-based procedural content generation~\cite{shaker2016procedural}.

There is another aspect of AI-assisted game design which we believe is hugely under-explored: automatic game tuning.  This involves taking an existing game (either human-designed or auto-generated) and performing a comprehensive exploration of the parameter space to find the most interesting game instances.

Recent work has demonstrated the potential of this approach, automatically generating distinct and novel variants of the minimalist mobile game \emph{Flappy Bird}~\cite{isaksen2015discovering}. 
That work involved using a very simple agent to play through each generated game instance.
Noise was added to the selected actions, and a game variant was deemed to have an appropriate level of difficulty if a specified number of players achieved a desired score. For Flappy Bird it is straightforward to design an AI agent capable of near-optimal play. Adding noise to the selected actions of this player can be used to provide a less than perfect agent that better represents human reactions. An evolutionary algorithm was used to search for game variants that were as far apart from each other in parameter space as possible but were still playable.

However, for more complex games it is harder to provide a good AI agent, and writing a new game playing agent for each new game would make the process more time consuming.  Furthermore,  a single hand-crafted agent may be blind to novel aspects of evolved game-play elements that the designer of the AI agent had not considered.  This could severely inhibit the utility of the approach.  In this work we mitigate these concerns by tapping in to an ever-growing pool of agents designed for the General Video Game AI (GVG-AI) competition\footnote{\url{http://www.gvgai.net/}}. The idea is that using a rich set of general agents will provide the basis for a robust evaluation process with a higher likelihood of finding skill-depth wherever it may lie in the chosen search space of possible games. In this paper we use one of the sample GVG-AI agents, varying it by changing the rollout budget. This was done by making the game implement a standard GVG-AI game interface, so that any GVG-AI agent can be used with very little effort, allowing the full set of agents to be used in future experiments.


Liu et al.~\cite{liu2016rolling} introduced a two-player space-battle game, derived from the original \emph{Spacewar}, and performed a study on different parameter settings to bring out some strengths and weaknesses of the various algorithms under test. A key finding is that the rankings of the algorithms depend very much on the details of the game. A mutation of one parameter may lead to a totally different ranking of algorithms. If the game using only a single parameter setting is tested, the conclusions could be less robust and misleading.

In this paper, we adapt the space-battle game introduced by Liu et al.~\cite{liu2016rolling} to the GVG-AI framework, then uses the Random Mutation Hill Climber (RMHC) and Multi-Armed Bandit RMHC (MABRMHC) to evolve game parameters to provide some game instances that lead to high winning rates for GVG-AI sample MCTS agents. This is used as an approximate measure of skill-depth, the idea being that the smarter MCTS agents should beat unintelligent agents, or that MCTS agents with a high rollout budget should beat those with a low rollout budget.

The paper is structured as follows: Section \ref{sec:agd} provides a brief review of the related work on automatic game design,
Section~\ref{sec:fram} describes the game engine, Section~\ref{sec:optimiser} introduces the two optimisation algorithms used in this paper,  Section~\ref{sec:xp} presents the experimental results, finally Section~\ref{sec:conc} concludes and discusses the potential directions in the future.
\section{Automatic game design and depth estimation}\label{sec:agd}

Attempts to automatically design complete games go back to Barney Pell, who generated rules for chess-like games~\cite{pell1992metagame}. It did not however become an active research topic until the late 2000's.

Togelius et al.~\cite{togelius2007towards} evolved racing tracks in a car racing game using a simple multi-objective evolutionary algorithm called \emph{Cascading Elitism}. The fitness functions attempted to capture various aspects of player experience, using a neural network model of the player. This can be seen as an early form of experience-driven procedural content generation~\cite{togelius2011search}, where game content is generated through search in content space using evolutionary computation or some other form of stochastic optimisation. Similar methods have since been used to generate many types of game content, such as particle systems for weapons in a space shooter~\cite{hastings2009automatic}, platform game levels~\cite{sorenson2010towards} or puzzles~\cite{ashlock2010automatic}. In most of these cases, the fitness functions measure some aspect of problem difficulty, with the assumption that good game content should not make the game too hard nor too easy.

While the research discussed above focuses on generating content for an existing game, there have been several attempts to use the search-based methods to generate new games by searching though spaces of game rules. Togelius and Schmidhuber~\cite{togelius2008experiment} used a simple hill-climber to generate single-player Pac-Man-like games given a restricted rule search space. The fitness function was based on learnability of the game, operationalised as the capacity of another machine learning algorithm to learn to play the game. 

This approach was taken further by Cook et al.~\cite{cook2011multi,cook2012aesthetic}, who used search-based methods to design rulesets, maps and object layouts in tandem for producing simple arcade games via a system called ANGELINA. Further iterations of this system include the automatic selection of media sources, such as images and resources, giving this work a unique flavour. In a similar vein, Browne and Maire~\cite{browne2010evolutionary} developed a system for automatic generation of board games; they also used evolutionary algorithms, and a complex fitness function based on data gathered from dozens of humans playing different board games. Browne's work is perhaps the only to result in a game of sufficient quality to be sold as a stand-alone product; this is partly a result of working in a constrained space of simple board games.

A very different approach to game generation was taken by Nelson and Mateas~\cite{nelson2007towards}, who use reasoning methods to create Wario Ware-style minigames out of verb-noun relations and common minigame design patterns. Conceptnet and Wordnet were used to find suitable roles for game objects. 

Quite recently, some authors have used search-based methods to optimise the parameters of a single game, while keeping both game rules and other parts of the game content constant. In the introduction we discussed the work of Isaksen et al. on generating playable \emph{Flappy Bird} variants~\cite{isaksen2015discovering}. Similarly, Powley et al.~\cite{powley2016automated} optimise the parameters of an abstract touch-based mobile game, showing that parameter changes to a single ruleset can give rise to what feels and plays like different games.

One of the more important properties of a game can be said to be its \emph{skill depth}, often just called \emph{depth}. This property is universally considered desirable by game designers, yet it is hard to define properly; some of the definitions build on the idea of a skill chain, where deeper games simply have more things that can be learned~\cite{lantz2017depth}. Various attempts have been made to algorithmically estimate depth and use it as a fitness function; some of the research discussed above can be said to embody an implicit notion of depth in their fitness functions. Relative Algorithm Performance Profiles (RAPP) is a more explicit attempt at game depth estimation; the basic idea is that in a deeper game, a better player get relatively better result than a poorer player. Therefore, we can use game-playing agents of different strengths to play the same game, and the bigger the difference in outcome the greater the depth~\cite{nielsen2015general}.

In this paper we use a form of RAPP to try to estimate the depth of variants of a simple two-player game. Using this measure as a fitness function, we optimise the parameters of this game to try to find deeper game variants, using two types of Random-Mutation Hill-Climber. The current work differs from the work discussed above both in the type of game used (two-player physics-based game), the search space (a multi-dimensional discrete space) and the optimisation method. In particular, compared to previous work by Isaksen et al, the current paper investigates a more complex game and uses a significantly more advanced agent, and also optimizes for skill-depth rather than difficulty. This work is, as far as we know, the first attempt to optimize skill-depth that has had good results.

\section{Framework}\label{sec:fram}
We adapt the two-player space-battle game introduced by Liu et al.~\cite{liu2016rolling} to the GVG-AI framework, then use RMHC and MABRMHC to evolve game parameters to provide some game instances that lead to high winning rate for GVG-AI sample MCTS agents. The main difference in the modified space-battle game used in this work is the introduction of weapon system. 
Each ship has the choice to fire a missile after its cooldown period has finished. 
From now on, we use the term ``game'' to refer to a game instance, i.e. a specific configuration of game parameters.

\paragraph{Spaceship} 
Each player/agent controllers a spaceship which has a maximal speed, $v_s$ units distance per game tick, and slows down over time. At each game tick, the player can choose to \emph{do nothing} or to make an action among \emph{\{RotateClockwise, RotateAnticlockwise, Thrust, Shoot\}}. A missile is launched while the \emph{Shoot} action is chosen and its cooldown period is finished, otherwise, no action will be taken (like \emph{do nothing}). The spaceship is affected by a random recoil force when launching a missile.
\paragraph{Missile}
A missile has a maximal speed, $v_m$ units distance per game tick, and vanishes into nothing after $30$ game tick. It never damages its mother ship.

Every spaceship has a radius of 20 pixels and every missile has a radius of 4 pixels in a layout of size 640*480.
\paragraph{Score}
Every time a player hits its opponent, it obtains $100$ points (reward).
Every time a player launches a missile, it is penalized by $c$ points (cost).
Given a game state $s$, the player $i\in \{1,2\}$ has a $score$ calculated by:
\begin{equation}\label{eq:score}
score(i) = 100 \times nb_k(i) - c \times nb_m(i),
\end{equation}
where $nb_k(i)$ is the number of lives subtracted from the opponent and $nb_m(i)$ indicates the number of launched missiles by player $i \in \{1,2\}$.

\paragraph{End condition}
A game ends after $500$ game ticks.
A player wins the game if it has higher score than its opponent after $500$ game ticks, and it's a loss of the other player. If both players have the same score, it's a draw.

\begin{figure}[hbt]
\centering
\includegraphics[width=.4\textwidth]{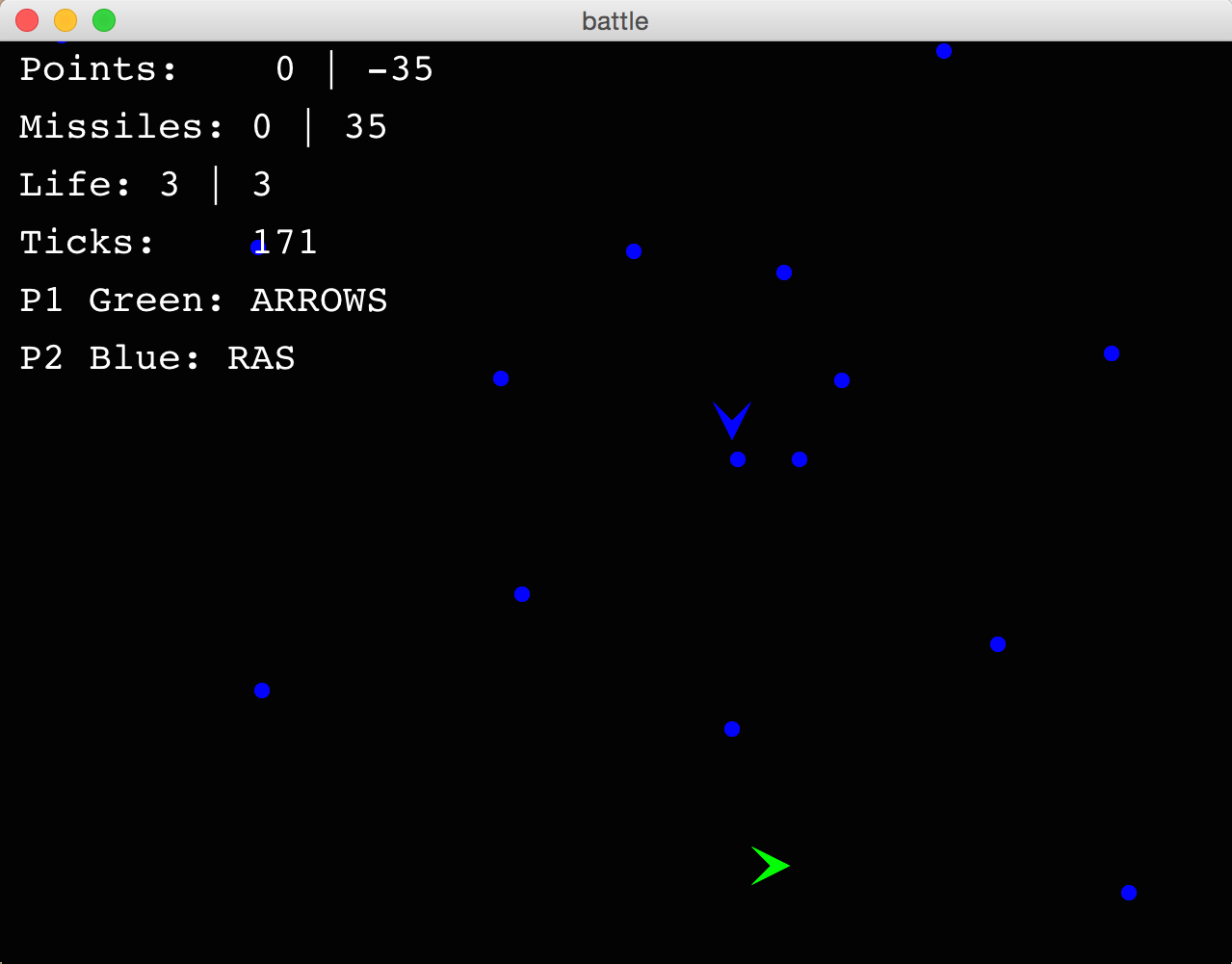}
\includegraphics[width=.4\textwidth]{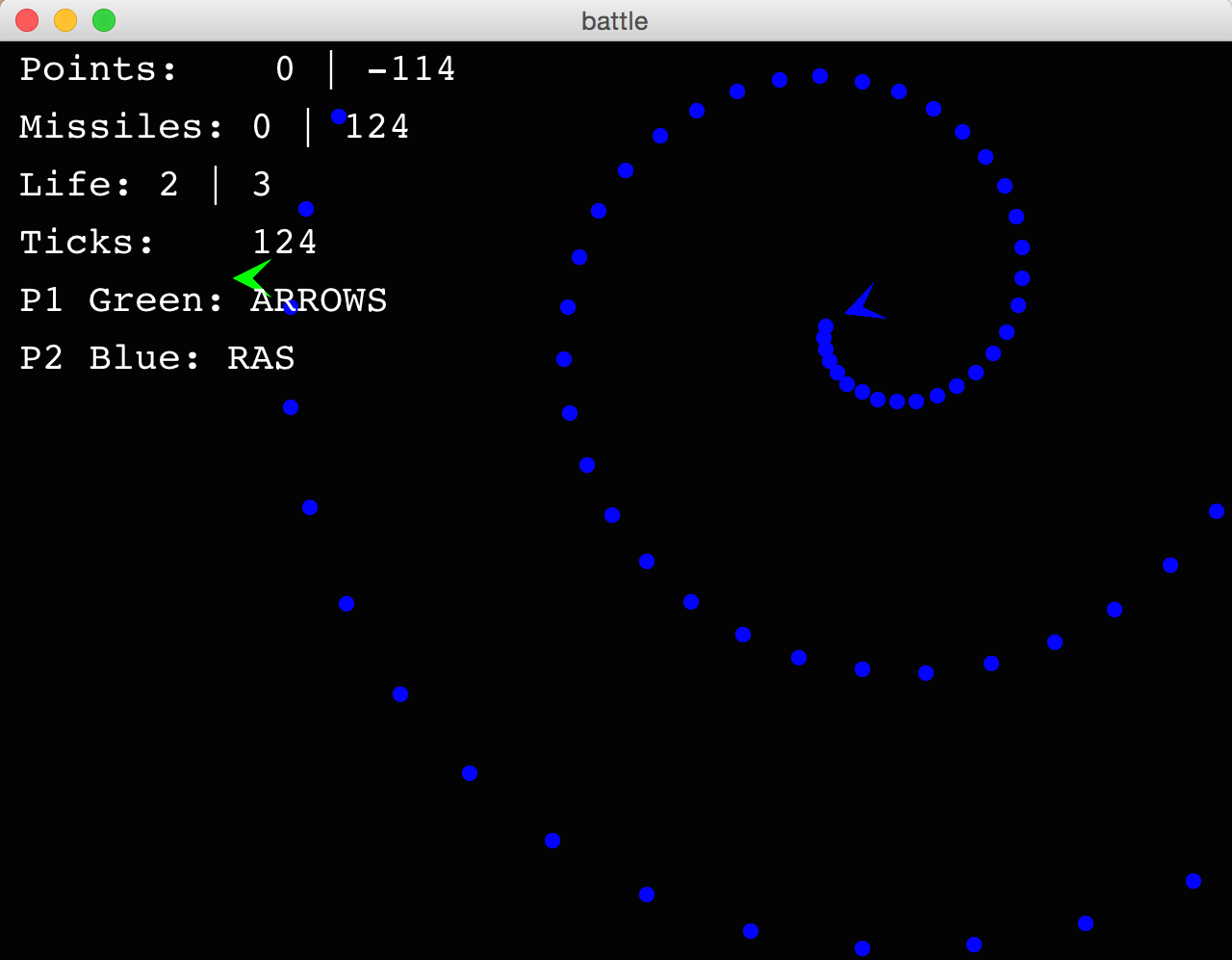}
\caption[ ]{\label{fig:ras}.Space-battle game with high (left) and low (right) missile cooldown time while fixing the other game parameters. It is more difficult to approach to RAS in the latter case.}
\end{figure}

\paragraph{Parameter space}\label{sec:space}
The parameters to be optimised are detailed in Table \ref{tab:params}. There are in total 14,400 possible games in the $5$-dimensional search space. Fig.~\ref{fig:ras} illustrates briefly how the game changes by varying only the cooldown time for firing missiles.
\begin{table}[h]
\centering
\caption{\label{tab:params}Game parameters. Only the first 5 parameters are optimised in the primary experiments. The last one (ship radius) is taken into account in Section \ref{sec:larger}.}
\scriptsize
\begin{tabular}{cccc}
\hline
Parameter & Notation & Legal values & Dimension\\
\hline
Maximal ship speed & $v_s$ & $4,6,8,10$ & 4 \\
Thrust speed & $v_t$ & $1,2,3,4,5$ & 5\\
Maximal missile speed & $v_m$ & $1,2,3,4,5,6,7,8,9,10$ & 10\\
Cooldown time & $d$ & $1,2,3,4,5,6,7,8,9$ & 9\\
Missile cost & $c$ & $0, 1, 5, 10, 20, 50, 75, 100$ & 8\\
Ship radius & $sr$ & $10, 20, 30, 40, 50$ & 5\\
\hline
\end{tabular}
\end{table}

The game is stochastic but fully observable. Each game starts with the agents in the symmetric positions. The two agents make simultaneous moves and in a fair situation. Thus, changing the player id does not change the situation of any player.
\section{Optimisers}\label{sec:optimiser}
We compare a Random Mutation Hill-Climber to an Multi-Armed Bandit Random Mutation Hill-Climber in evolving instances for space-battle game described previously.
This section is organised as follows.
Section \ref{sec:rmhc} briefly recalls the Random Mutation Hill-Climber. Section \ref{sec:mabrmhc} presents the Multi-Armed Bandit Random Mutation Hill-Climber and its selection and mutation rules.

\subsection{Random Mutation Hill-Climber}\label{sec:rmhc} 
The Random Mutation Hill-Climber (RMHC) is a simple but efficient derivative-free optimisation method mostly used in discrete domains~\cite{lucas2003learning,lucas2005learning}. 
The pseudo-code of RMHC is given in Algorithm \ref{algo:rmhc}. 
At each generation, an offspring is generated based on the only best-so-far genome (parent) by mutating exactly one uniformly randomly chosen gene.
The best-so-far genome is updated if the offspring's fitness value is better or equivalent to the best-so-far.
\begin{algorithm}[hbt]
\caption{\label{algo:rmhc}Random Mutation Hill-Climber (RMHC).}
\begin{algorithmic}[1]
\Require{$\mathcal{X}$: search space}
\Require{$D =|\mathcal{X}|$: problem dimension (genome length)}
\Require{$f: \mathcal{X} \mapsto [0,1]$: fitness function}
\State{Randomly initialise a genome $\mathbf{x} \in \mathcal{X}$}
\State{$bestFitSoFar \gets 0$}
\State{$M \gets 0$}	\Comment{Counter for the latest best-so-far genome}
\State{$N \gets 0$}	\Comment{Total evaluation count so far}
\While{time not elapsed}
   \State{Uniformly randomly select $d \in \{1,\dots,D\}$}
   \State{$\mathbf{y} \gets$ new genome by uniformly randomly mutating the $d^{th}$ gene of $\mathbf{x}$}
    \State{$Fit_{\mathbf{x}} \gets fitness(\mathbf{x})$} 
    \State{$Fit_{\mathbf{y}} \gets fitness(\mathbf{y})$}
    \State{$averageFitness_{\mathbf{x}} \gets \frac{bestFitSoFar\times M+Fit_{\mathbf{x}}}{M+1}$}
    \State{$N \gets N + 2$}	\Comment{Update evaluation count}
    \If{$Fit_{\mathbf{y}} \geq averageFitness_{\mathbf{x}}$}
    	\State{$\mathbf{x} \gets \mathbf{y}$}	\Comment{Replace the best-so-far genome}
   		\State{$bestFitSoFar \gets Fit_{\mathbf{y}}$}
        \State{$M \gets 1$}
    \Else
    	\State{$bestFitSoFar \gets averageFitness_{\mathbf{x}}$} 
        \State{$M \gets M + 1$}
    \EndIf
\EndWhile
\State{\Return{$\mathbf{x}$}}
\end{algorithmic}
\end{algorithm}

\subsection{Multi-Armed Bandit Random Mutation Hill-Climber}\label{sec:mabrmhc} 

Multi-Armed Bandit Random Mutation Hill-Climber (MABRMHC), derived from the 2-armed bandit-based RMHC \cite{liu2016bandit,liu2017banditrmhc}, uses both UCB-style selection and mutation rules. MABRMHC selects the coordinate (bandit) with the maximal $urgency$ (Equation \ref{eq:UrgentUCB}) to mutate, then mutates the parameter in dimension $d$ to the value (arm) which leads to the maximal reward (Equation \ref{eq:ucb}).

For any multi-armed bandit $d\in\{1,2,\dots,D\}$, its $urgency_d$ is defined as
\begin{equation} \label{eq:UrgentUCB}
\begin{aligned}
&urgency_d= \\
&\underset{1\leq j \leq Dim(d)}{\min} \left(\Delta_d(j) + \sqrt{\frac{2\log(\sum_{k=1}^{Dim(d)}N_d(k))}{N_d}} + \w\right),
\end{aligned}
\end{equation}
where $N_d(k)$ is the number of times the $k^{th}$ value is selected when the $d^{th}$ coordinate is selected; $N_d$ is the number of times the $d^{th}$ coordinate is selected to mutate, thus $N_d=\sum_{k=1}^{Dim(d)}N_d(k)$; $\Delta_d(k)$ is the maximal difference between the fitness values if the value $k$ is mutated to when the $d^{th}$ dimension is selected, i.e., the changing of fitness value; $\w$ denotes a uniformly distributed value between $0$ and $1e^{-6}$ which is used to randomly break ties.
Once the coordinate to mutate (eg. $d^*$) is selected, the index of the value to mutated to is determined by
\begin{equation}\label{eq:ucb}
k^*=\underset{1\leq k\leq Dim(d^*)}{\argmax} \left(  \bar\Delta_{d^*}(k) + \sqrt{\frac{2\log(N_{d^*})}{N_{d^*}(k)}} + \w\right),
\end{equation}
where $\bar\Delta_{d^*}(k)$ denotes the average changing of fitness value if the value $k$ is mutated to when the dimension $d^*$ is selected.

The pseudo-code of MABRMHC is given in Algorithm \ref{algo:mabrmhc}.

\begin{algorithm}[hbtp]
\caption{\label{algo:mabrmhc}Multi-Armed Bandit Random Mutation Hill-Climber (MABRMHC). $Dim(d)$ returns the number of possible values in dimension $d$. $\w$ denotes a uniformly distributed value between $0$ and $1e^{-6}$ which is used to randomly break ties.}
\begin{algorithmic}[1]
\Require{$\mathcal{X}$: search space}
\Require{$D =|\mathcal{X}|$: problem dimension (genome length)}
\Require{$f: \mathcal{X} \mapsto [0,1]$: fitness function}
\State{Randomly initialise a genome $\mathbf{x} \in \mathcal{X}$}
\State{$bestFitSoFar \gets 0$}
\State{$M \gets 0$}	\Comment{Counter for the latest best-so-far genome}
\State{$N \gets 0$}	\Comment{Total evaluation count so far}
\For{$d \in \{1,\dots,D\}$}
\State{$N_d=0$}
\For{$k \in \{1,\dots,Dim(d)\}$}
\State{$N_d(k)=0$, $\Delta_{d}(k)=0$, $\bar\Delta_{d}(k)=0$}
\EndFor
\EndFor
\While{time not elapsed}
	\State{$d^*=\underset{1\leq d\leq  D}{\argmax} \left( \underset{1\leq j \leq Dim(d)}{\min} \Delta_d(j) + \sqrt{\frac{2\log(\sum_{k=1}^{Dim(d)}N_d(k))}{N_d}} + \w\right)$} \Comment{Select the coordinate to mutate (Equation \ref{eq:UrgentUCB})}
    \State{$k^*=\underset{1\leq k\leq Dim(d^*)}{\argmax} \left(  \bar\Delta_{d^*}(k) + \sqrt{\frac{2\log(N_{d^*})}{N_{d^*}(k)}} + \w\right)$} \Comment{Select the index of value to take (Equation \ref{eq:ucb})}
    \State{$\mathbf{y} \gets$ after mutating the element $d^*$ of $\mathbf{x}$ to the $k^*$ legal value}     
    \State{$Fit_{\mathbf{x}} \gets fitness(\mathbf{x})$}
    \State{$Fit_{\mathbf{y}} \gets fitness(\mathbf{y})$}
    \State{$averageFitness \gets \frac{bestFitSoFar\times M+Fit_{\mathbf{x}}}{M+1}$}
    \State{$N \gets N + 2$}	\Comment{Update the counter}
    \State{$\Delta = Fit_{\mathbf{y}} - averageFitness$}
    \State{Update $\Delta_{d^*}(k^*)$ and $\bar\Delta_{d^*}(k^*)$} \Comment{Update the statistic}
    \State{$N_d(k)\gets N_d(k)+1$, $N_d\gets N_d +1$}\Comment{Update the counters}
    \If{$\Delta \geq 0$}
    	\State{$\mathbf{x} \gets \mathbf{y}$}	\Comment{Replace the best-so-far genome}
   		\State{$bestFitSoFar \gets Fit_{\mathbf{y}}$}
        \State{$M \gets 1$}
    \Else
    	\State{$bestFitSoFar \gets averageFitness$}
        \State{$M \gets M + 1$}
    \EndIf
\EndWhile
\State{\Return{$\mathbf{x}$}}
\end{algorithmic}
\end{algorithm}

In this work, we model each of the game parameter to optimise as a bandit, and the legal values for the parameter as the arms of this bandit. The search space is folded in the sense that it takes far less computational cost to mutate and evaluate every legal value of each parameter once than to evaluate mutate and evaluate every legal game instance once.

\section{Experimental results}\label{sec:xp}
We firstly use the sample agent using a two-player Open-Loop Monte-Carlo Tree Search algorithm provided by the GVG-AI framework, which uses the difference of scores (Eq. 
\ref{eq:score}) of both players as its heuristic (denoted as OLMCTS), as player 1. No modification or tuning has been performed on this sample agent. We implement a consistently rotate-and-shoot agent (denoted as RAS) as the player 2. More precisely, the RAS is a deterministic agent and, by Eq. \ref{eq:score}, the OLMCTS aims at maximising $\left(100 \times nb_k(1) - c \times nb_m(1)\right) - \left(100 \times nb_k(2) - c \times nb_m(2)\right)$,
where $nb_k(1)$ and $nb_k(2)$ are the numbers of lives subtracted from the RAS and OLMCTS, respectively; $nb_m(1)$ and $nb_m(2)$ indicates the number of launched missiles by OLMCTS and RAS, respectively. Again, this heuristic is already defined in the sample agent, not by us.
Basically, a human player could probably choose to play the game in a passive way by avoiding the missiles and not firing at all, and finally win the game.

The landscape of winning rate of OLMCTS against RAS is studied in Section \ref{sec:land}.
Section \ref{sec:optimise} presents the performance of RMHC and MABRMHC with different resampling numbers to generate games in the parameter space detailed previously (Section \ref{sec:space}) and Section \ref{sec:larger} presents their performances in a 5 times larger parameter space.

\subsection{Winning rate distribution}\label{sec:land}
We use a OLMCTS agent as the player 1 and a RAS agent as the player 2. At each game tick, $10ms$ is allocated to each of the agents to decide an action. The average number of iterations performed by OLMCTS is $350$. The time to return an action for RAS is negligible.

\begin{figure}[hbtp]
\centering
\includegraphics[width=.49\textwidth]{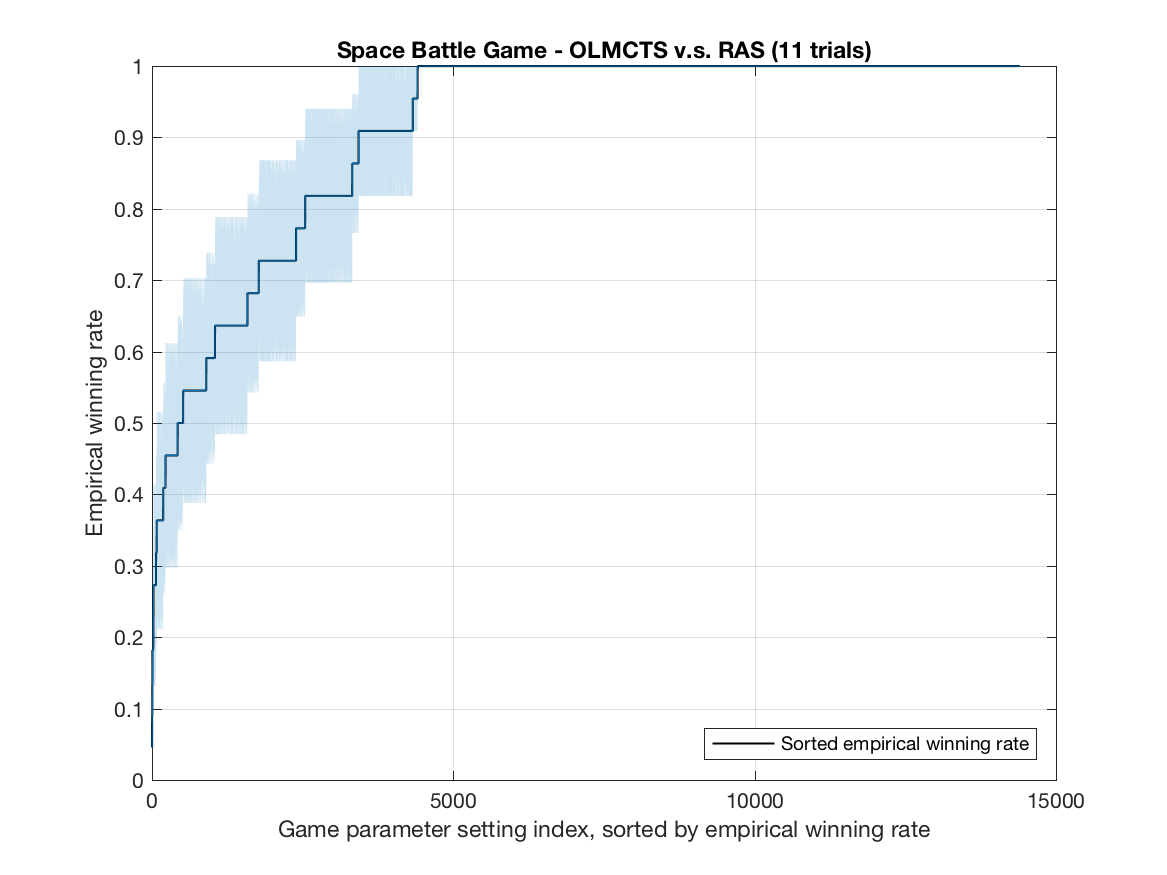}
\includegraphics[width=.49\textwidth]{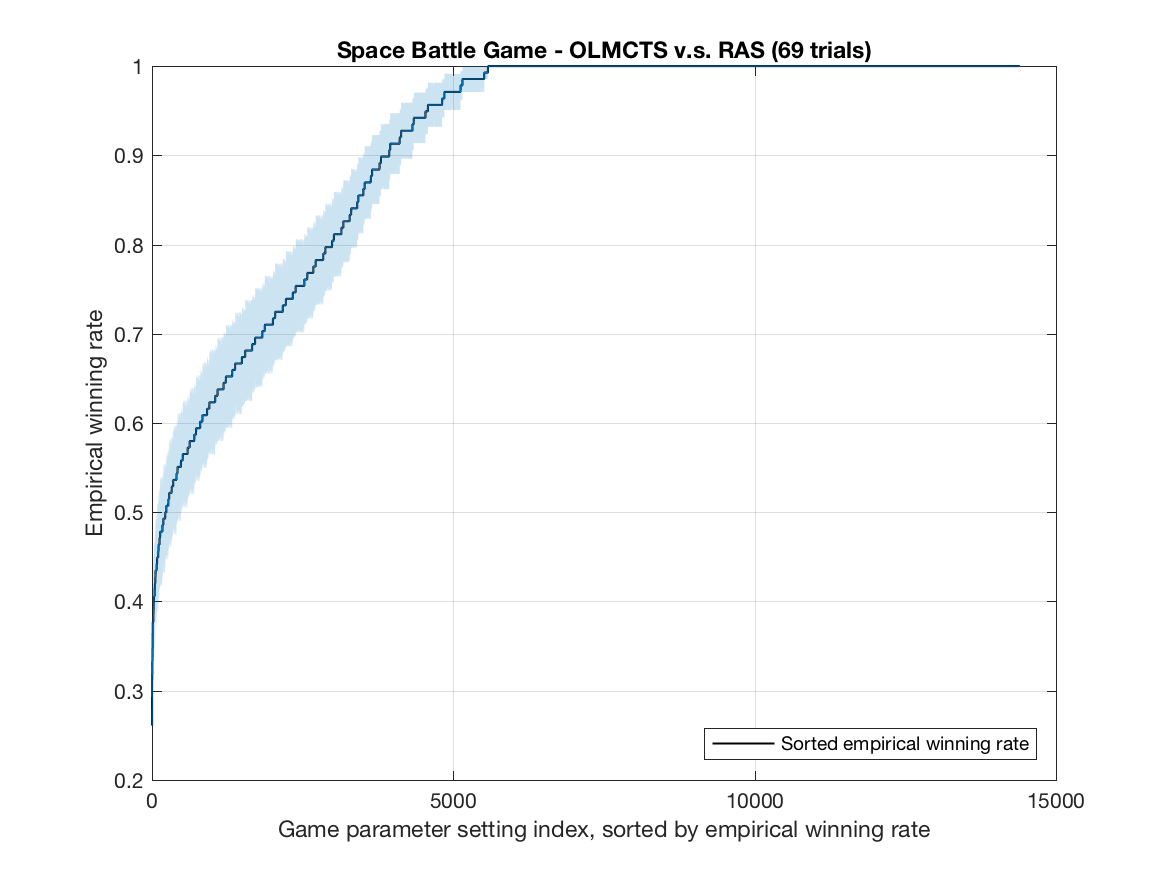}
\caption[ ]{\label{fig:wr}Empirical winning rates for OLMCTS sorted in increasing order, over 11 trials (left) and 69 trials (right), of all the 14,400 legal game instances played by OLMCTS against RAS. The standard error is shown by the shaded boundary.}
\end{figure}

The average winning rates over 11 and 69 repeated trials of all the 14,400 legal game instances played by OLMCTS against RAS are shown in Fig.
\ref{fig:wr}.
The winning rate over 69 trials of each games instance varies between 20\% and 100\%. Among all the legal game instances, the OLMCTS does not achieve a 100\% winning rate in more than 5,000 games. 

\begin{figure*}[hbt]
\centering
\includegraphics[width=.32\textwidth]{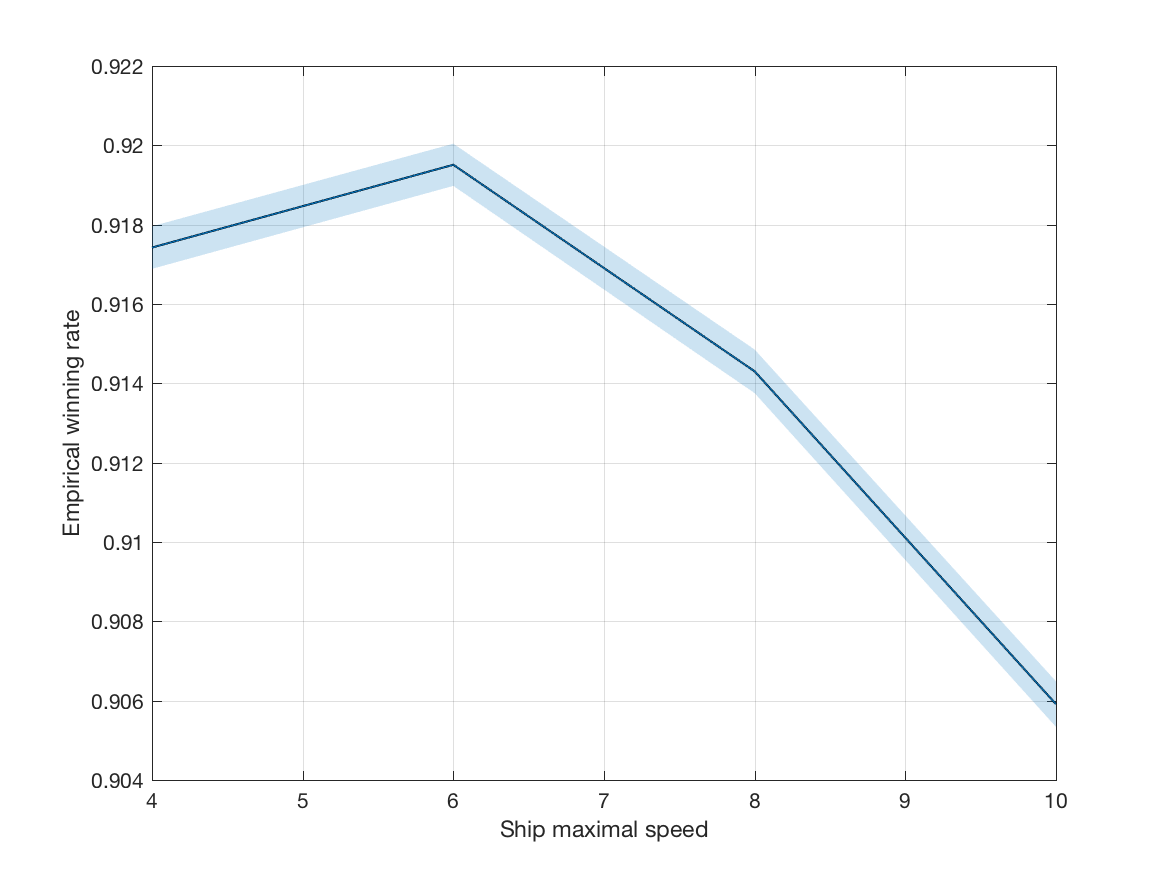}
\includegraphics[width=.32\textwidth]{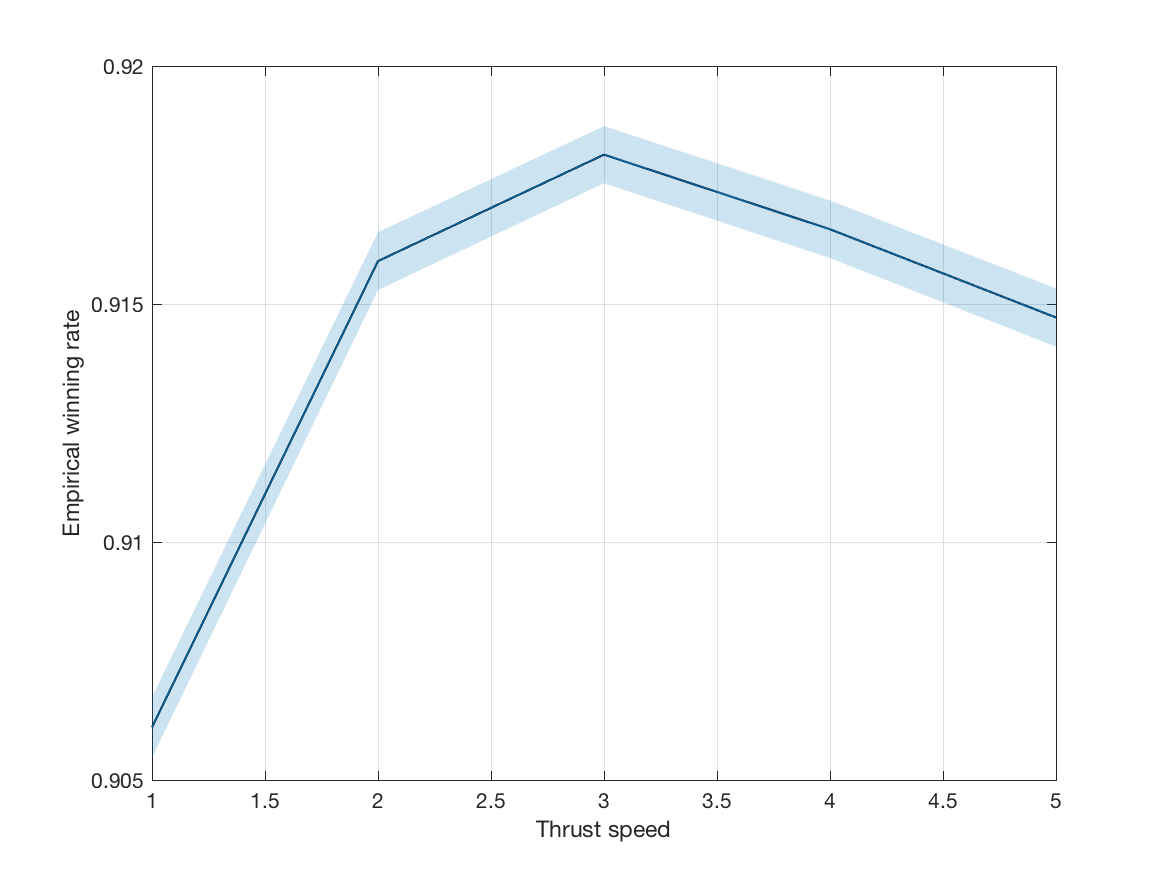}\\
\includegraphics[width=.32\textwidth]{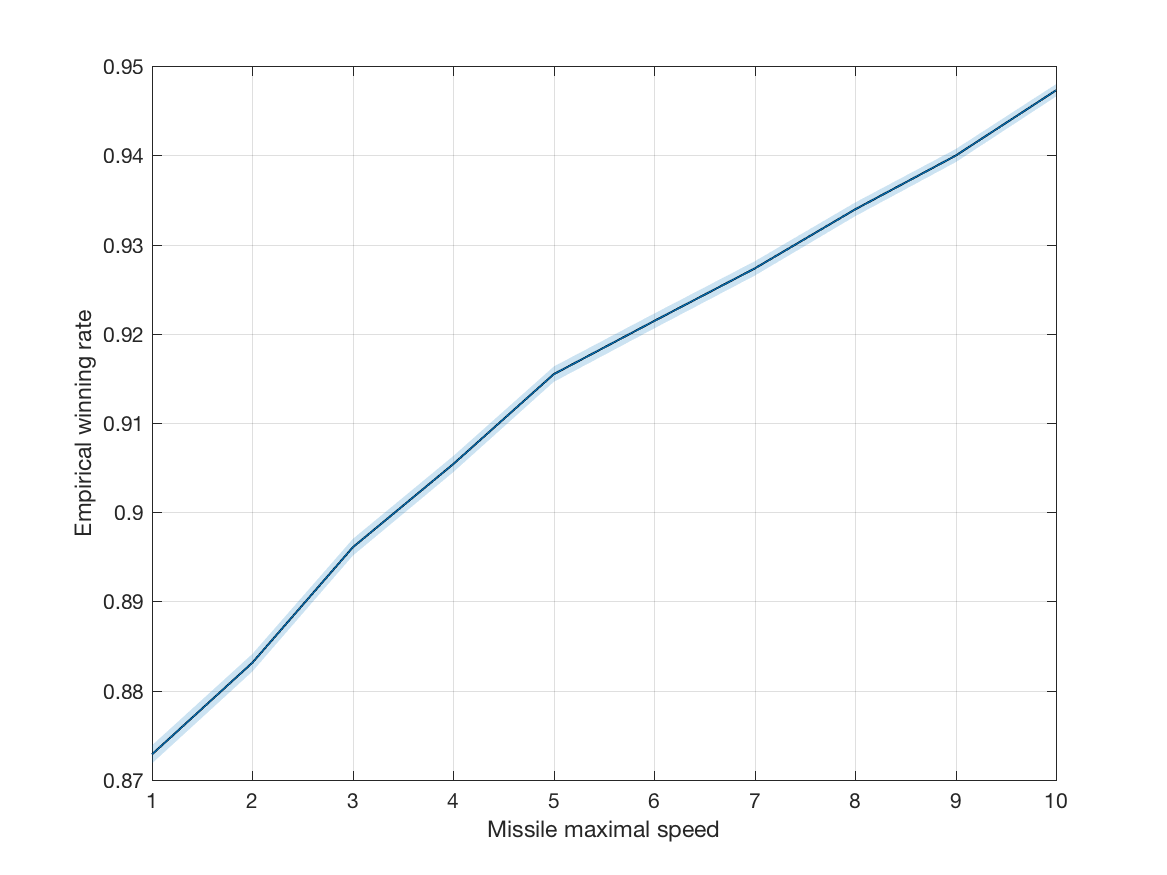}
\includegraphics[width=.32\textwidth]{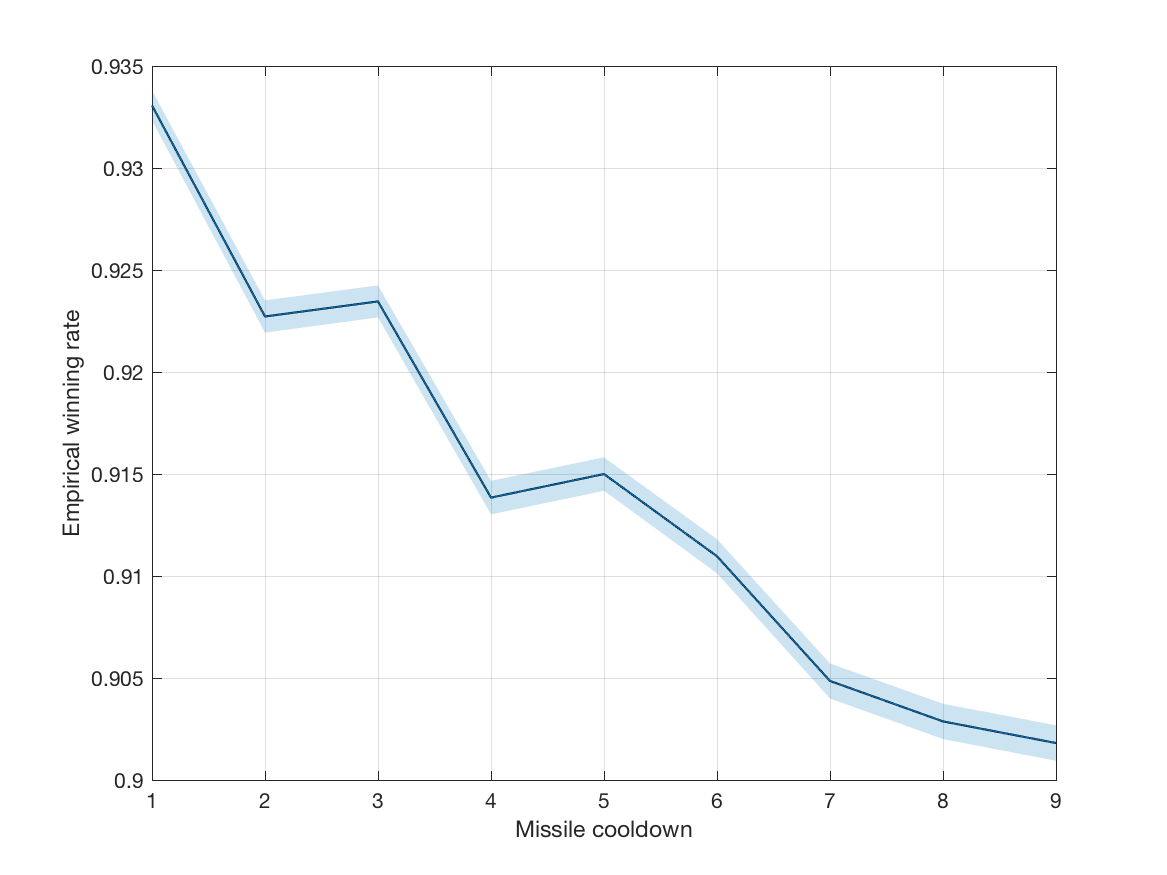}
\includegraphics[width=.32\textwidth]{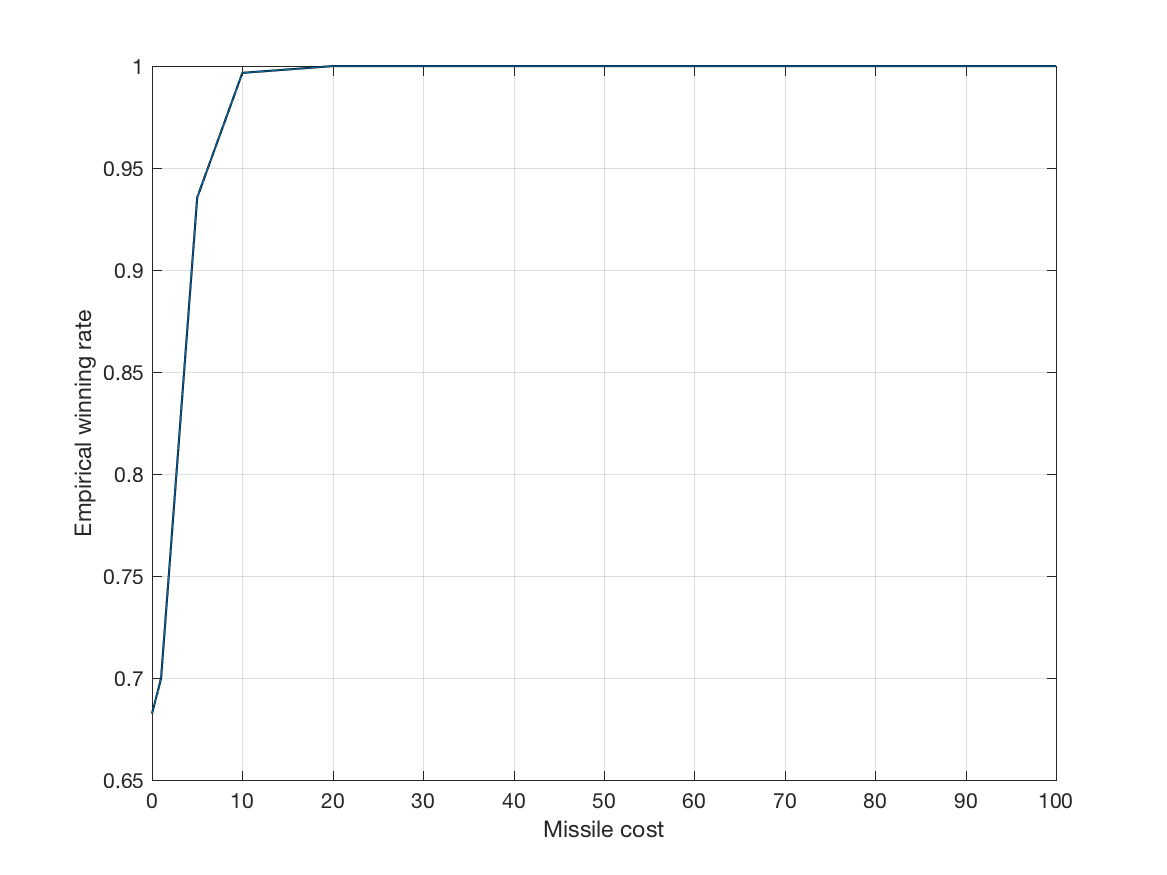}
\caption[ ]{\label{fig:ms}
Empirical winning rates over 69 trials of all the 14,400 legal game instances played by OLMCTS against RAS, classified by the maximal ship  speed, the thrust speed, the maximal missile speed, the cooldown time and the cost of firing a missile, respectively. The standard error is shown by the shaded boundary.}
\end{figure*}

Fig. \ref{fig:ms} demonstrates how the winning rate varies along with the changing of each parameter.
The maximal ship speed and the thrust speed have negligible influence on the OLMCTS's average winning rate. Higher the maximal missile speed is or shorter the cooldown time is, higher the average winning rate is. But still, the average winning rate remains above 87\%. The most important factor is the cost of firing a missile. It is not surprising, since the RAS fires successively missiles and the number of missiles it fires during each game is constant depending on the cooldown time. the OLMCTS only fires while necessary or it is likely to slash its opponent.

\subsection{Evolving games by RMHC and MABRMHC using different resampling numbers}\label{sec:optimise}
We use the same agents as described in Section \ref{sec:land}.
RMHC (Algorithm \ref{algo:rmhc}) and MABRMHC (Algorithm \ref{algo:mabrmhc}) are applied to optimise the parameters of the space-battle game, aiming at maximising the win probability for the OLMCTS against the RAS. 
Since the true win probability is unknown, we need to define the fitness of a game using some winning rate by repeating the same game several times, i.e., resampling the game.
We define the fitness value of a game $g$ as the winning rate over $r$ repeated games, i.e.,
\begin{equation}
fitness(g)= \frac{1}{r} \sum_{i=1}^{r} GameValue(g).\label{eq:fitness}
\end{equation}
The value of game $g$ is defined as
\[
    GameValue(g)= 
\begin{cases}
    1,& \text{if OLMCTS wins}\\
    0,& \text{if RAS wins}\\
    0.5,& \text{otherwise (a draw).}
\end{cases}
\]
A call to $fitness(\cdot)$ is actually based on independent $r$ realizations of the same game. Due to the internal stochastic effects in the game, each realization may return a different game value.
We aim at maximising the fitness $f$ in this work.
The empirical winning rates shown in Fig.~\ref{fig:wr} are two example $fitness(\cdot)$ with $r=11$ (left) and $r=69$ (right). The strength of noise decreases while repeating the same game more times, i.e., increasing $r$.

A recent work applied the RMHC and a two-armed bandit-based RMHC with resamplings to a noisy variant of the OneMax problem, and showed both theoretically and practically the importance of choosing a suitable resampling number to accelerate the convergence to the optimum~\cite{liu2016optimal,liu2016bandit,liu2017banditrmhc}. As the space-battle game introduced previously is stochastic and the agents can be stochastic as well, it is not trivial to model the noise or provide mathematically any optimal resampling number. Therefore, in this work, some resampling numbers are arbitrarily chosen and compared to give a primary idea about the necessary number of resamplings.

\begin{figure*}[h]
\centering
\includegraphics[width=.49\textwidth]{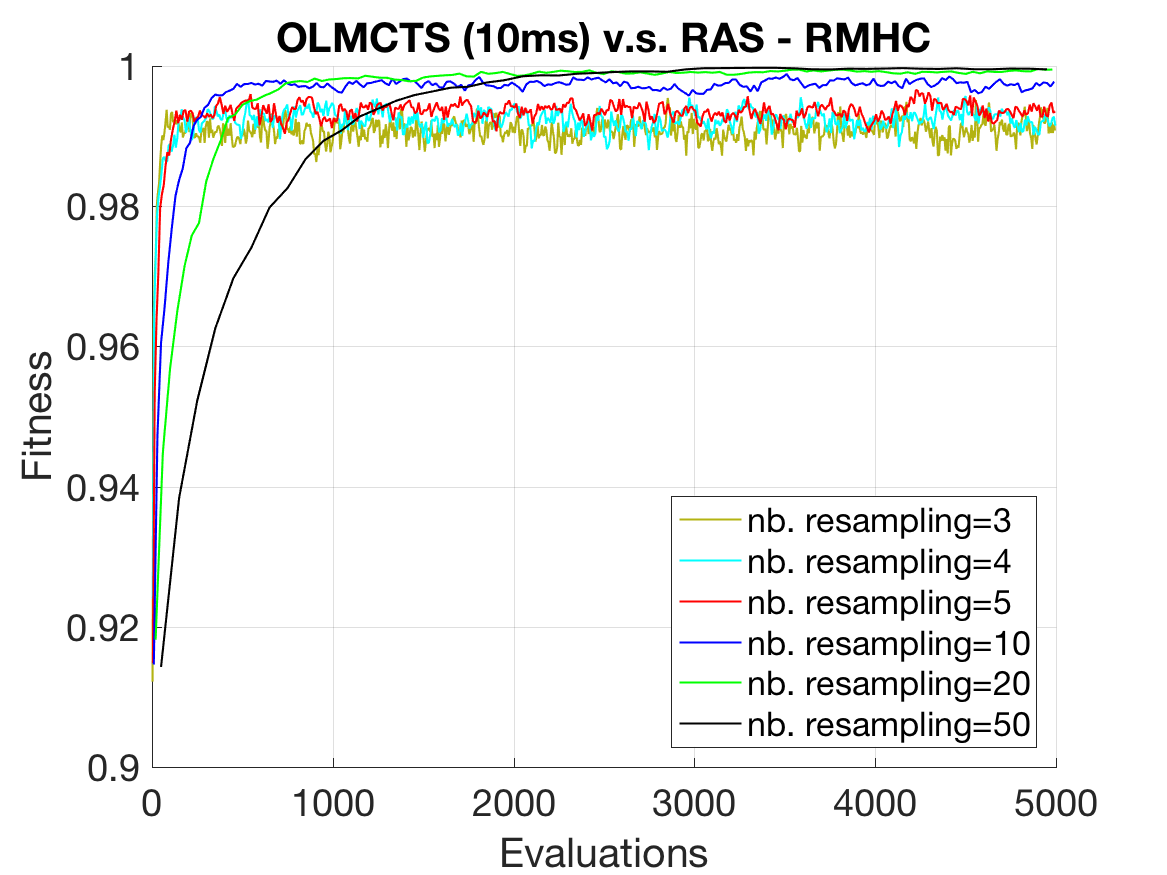}
\includegraphics[width=.49\textwidth]{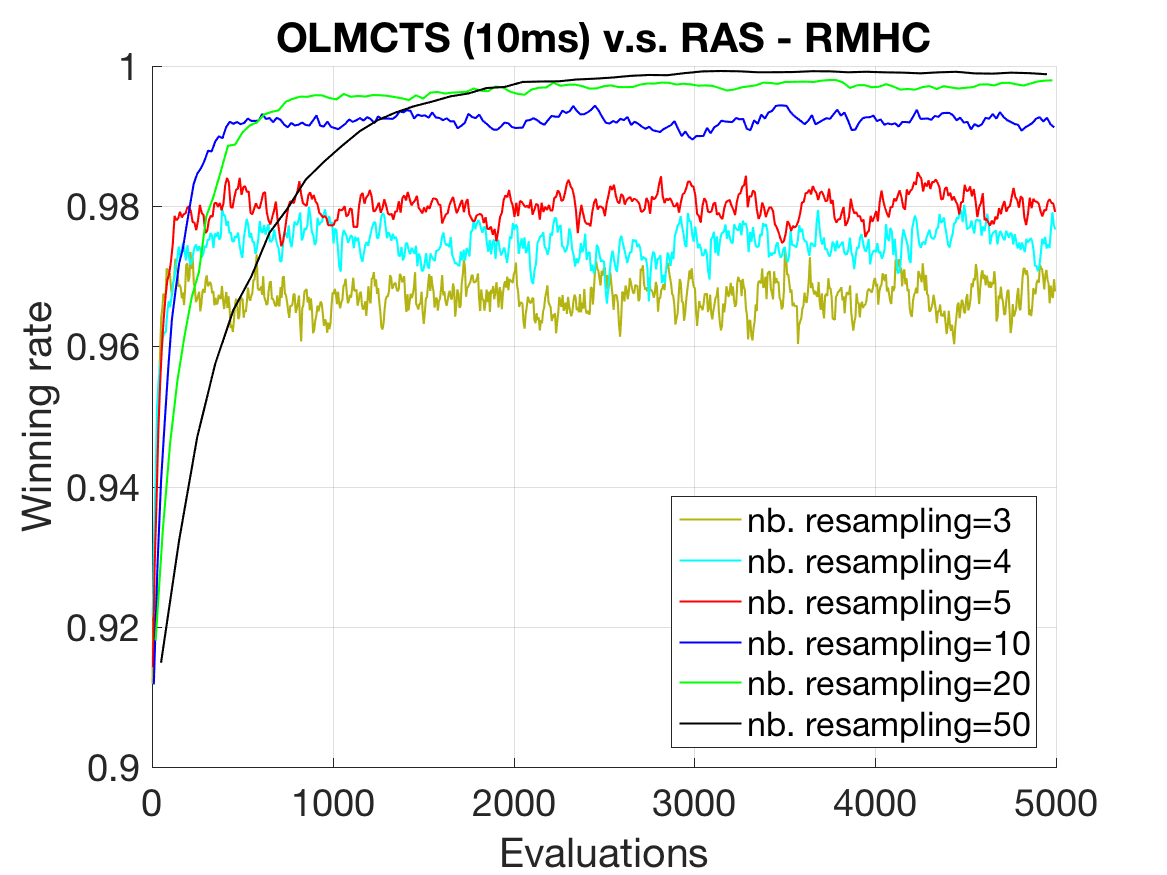}
\centering
\caption[ ]{\label{fig:mabrmhc}Average fitness value (left) with respect to the evaluation number over 1,000 optimisation trials by RMHC. The average winning rate of recommended game instances at each generation are shown on the right. The standard error is shown by the shaded boundary.}
\includegraphics[width=.49\textwidth]{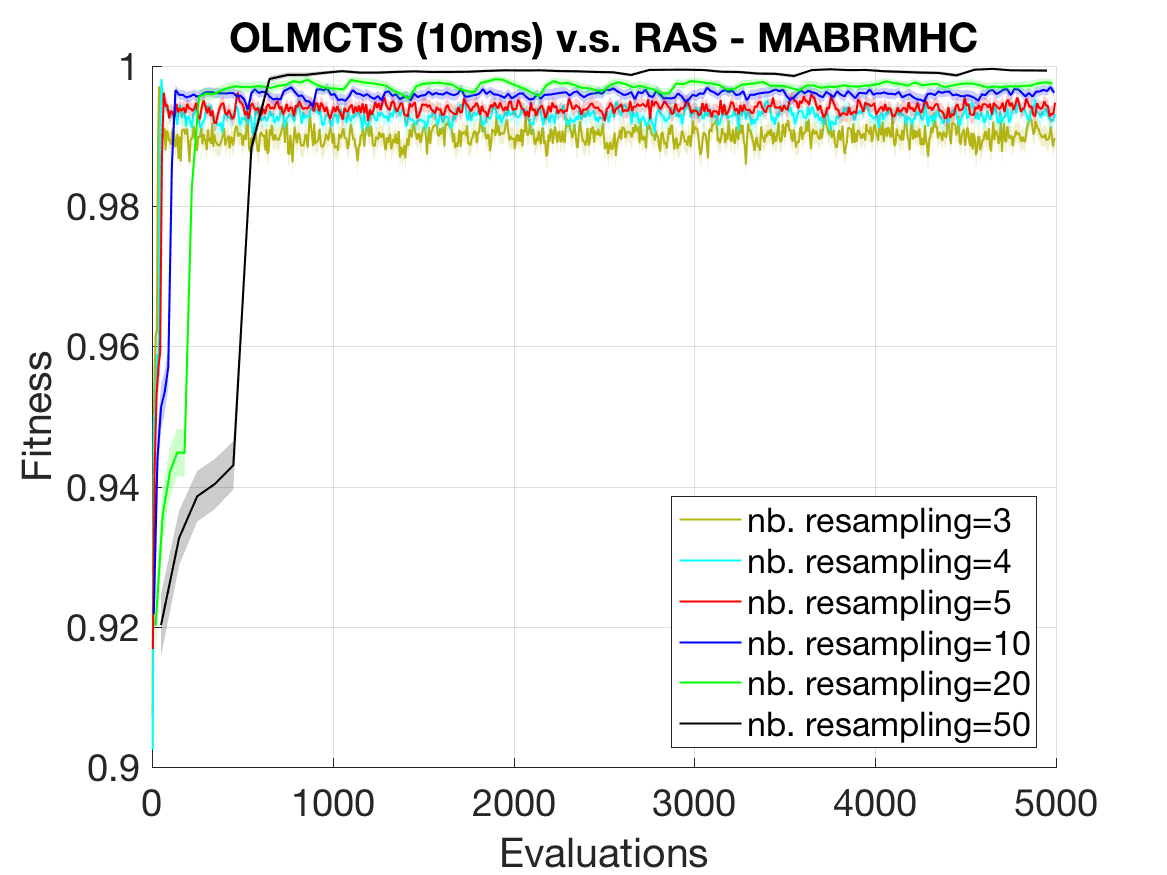}
\includegraphics[width=.49\textwidth]{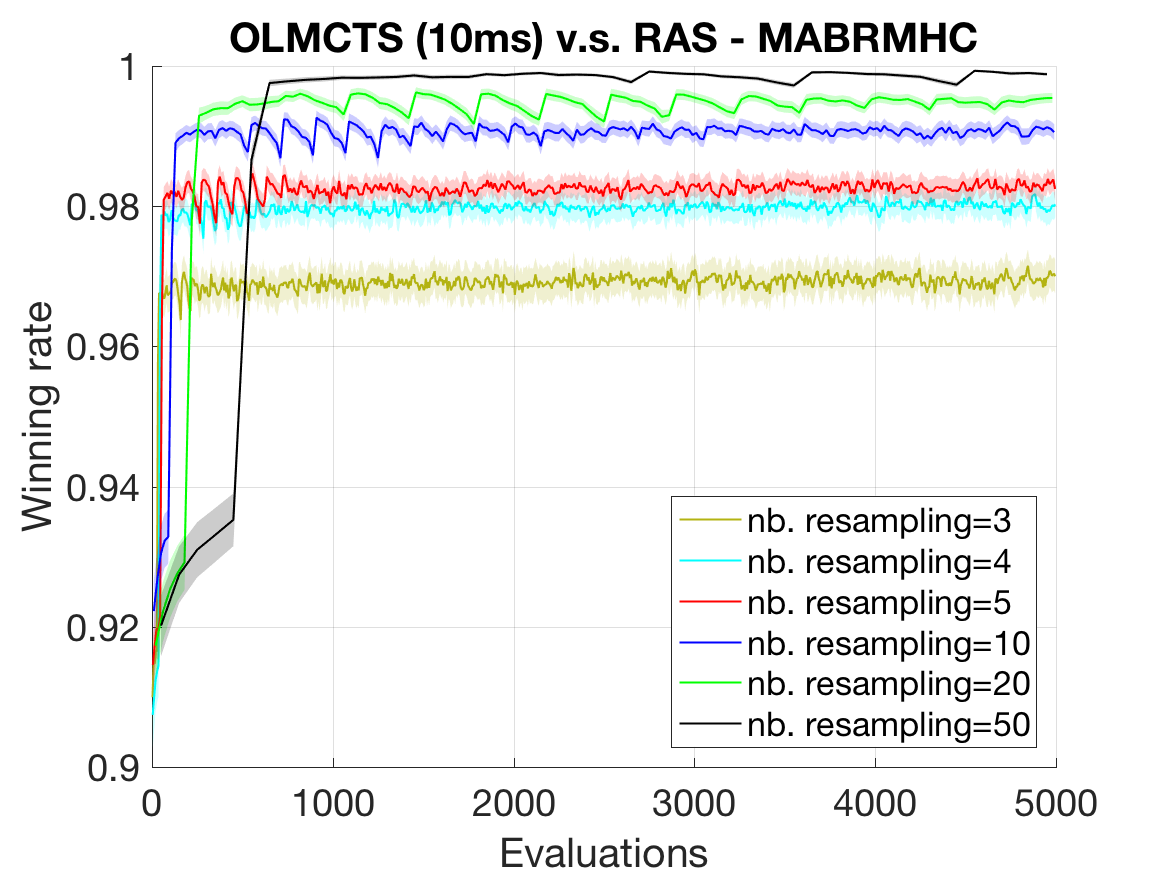}
\caption[ ]{\label{fig:rmhc}Average fitness value (left) respected to the evaluation number over 1,000 optimisation trials by MABRMHC. The average winning rate of recommended game instances at each generation are shown on the right. The standard error is shown by the shaded boundary.}
\end{figure*}

Figs. \ref{fig:rmhc} and \ref{fig:mabrmhc} illustrate the overall performance of RMHC and MABRMHC using different resampling numbers over 1,000 optimisation trials with random starting parameters. A number of 5,000 game evaluations is allocated as optimisation budget in each trial. In other words, given a resampling number $r$, the $fitness(\cdot)$ (Eq. \ref{eq:fitness}) is called at most $5,000/r$ times.

RMHC and MABRMHC using smaller resampling number achieve a faster move towards to the neighborhood of the optimum at the beginning of optimisation, however, they do not converge to the optimum along with time; despite the slow speed at the beginning, RMHC and MABRMHC using larger resampling number finally succeed in converging to the optimum in the limited budget. 
A dynamic resampling number which \emph{smoothly} increases with the number of generations will be favourable. 

Using smaller budget, MABRMHC reaches the neighborhood of the optimum faster than RMHC. While the current best-so-far fitness is near the optimal fitness value, it's not surprising to see the jagged curves (Fig. \ref{fig:mabrmhc}, right) while the game evaluation consumed is moderate. The drop to the valley dues to the exploration of MABRMHC, then it manages to return to the previous optimum found or possibly find another optimum. Along with the increment of budget, i.e., game evaluations, the quality of best-so-far games found by MABRMHC remains stable.

\subsection{Evolving games in a larger search space}\label{sec:larger}
All the 5 parameters considered previously are used for evolving the game rules. In this section, we expand the parameter space by taking into account a parameter for graphical object: the radius of ship. The legal values for ship's radius are $10, 20, 30, 40$ and $50$. Thus, the search space is $6$-dimensional and the total number of possible games is increase to 72,000 (5 times larger).

Instead of an intelligent agent and a deterministic agent, we play the same OLMCTS agent (with 350 iterations), which has been used previously in Section \ref{sec:land} and Section \ref{sec:optimise}, against two of its instances: a OLMCTS with 700 iterations and a OLMCTS with 175 iterations, denoted as OLMCTS700 and OLMCTS175 respectively. The same optimisation process using RMHC and MABRMHC is repeated separately, using 1,000 game evaluation. The resampling numbers used are 5 and 50, the ones which have achieved either fastest convergence at the beginning or provides the best recommendation at the end of optimisation (after 1,000 game evaluations), respectively.
We aim at verifying if the same algorithms still perform well in a larger parameter space and with smaller optimisation budget.
\begin{figure*}[h]
\centering
\includegraphics[width=.49\textwidth]{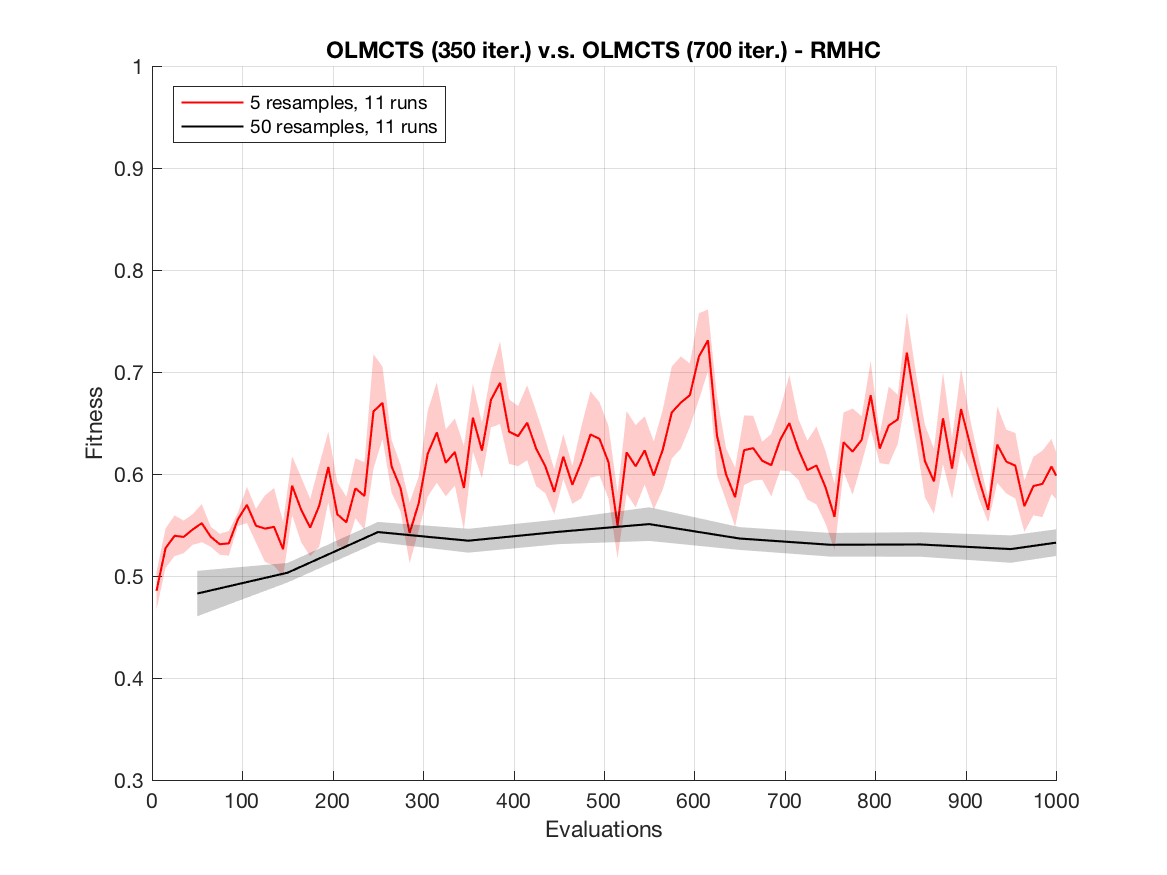}
\includegraphics[width=.49\textwidth]{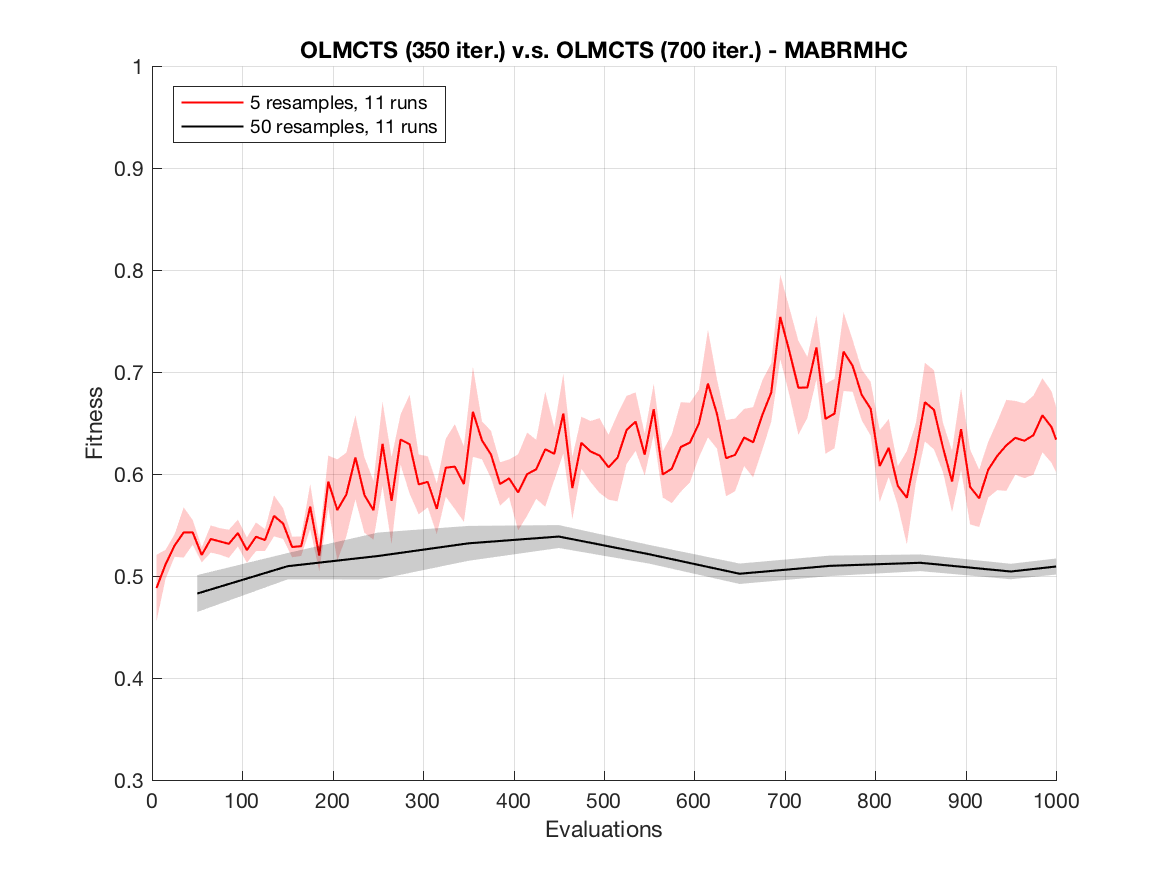}
\includegraphics[width=.49\textwidth]{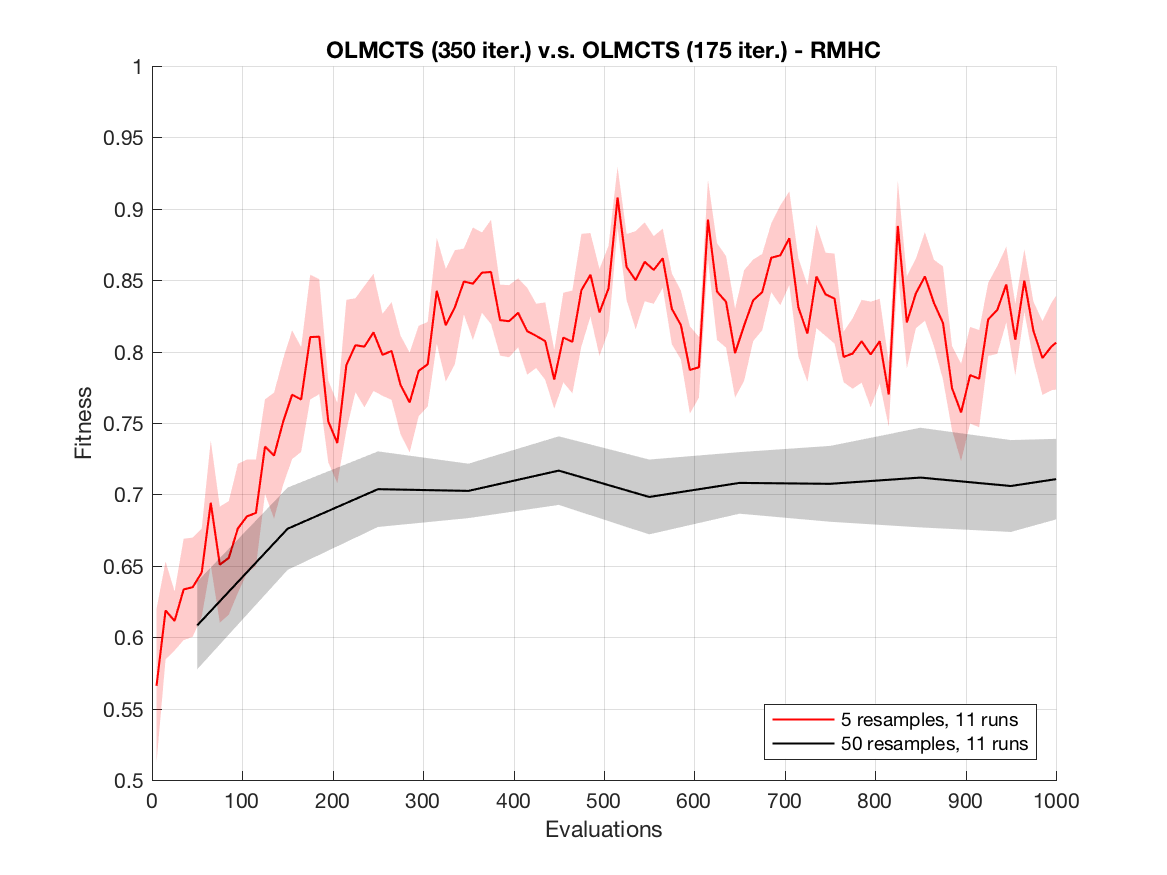}
\includegraphics[width=.49\textwidth]{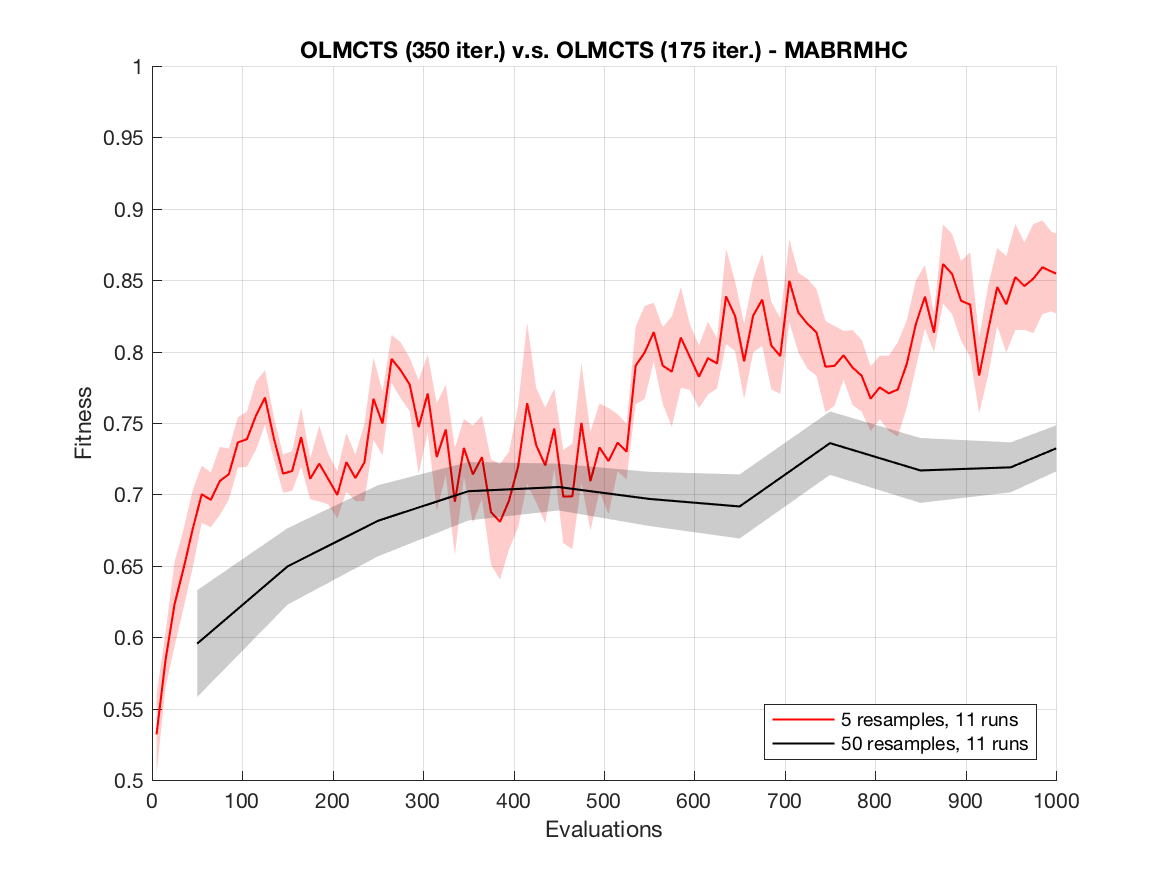}
\caption[ ]{\label{fig:lessrmhc}Average fitness value respected to the evaluation number over 11 optimisation trials by RMHC (left) and MABRMHC (right) using different resampling numbers. The games are played by OLMCTS with 350 iterations against OLMCTS with 700 iterations (top) or OLMCTS against OLMCTS with 175 iterations (bottom). The standard error is shown by the shaded boundary.}
\end{figure*}

Fig. \ref{fig:lessrmhc} shows the average fitness value respected to the number of game evaluations over 11 optimisation trials.
Resampling 50 times (black curves in Fig. \ref{fig:lessrmhc}) the same game instance guarantee a more accurate winning rate, while resampling 5 times (red curves in Fig. \ref{fig:lessrmhc}) seems to converge faster.

To validate the quality of recommendations, we play each recommended game instance, optimised by playing OLMCTS against OLMCTS175, 100 times using the OLMCTS175 and a random agent, which uniformly randomly returns a legal action. The idea is to verify that the game instances optimised for OLMCTS, are still playable and beneficial for OLMCTS instance with small number of iterations.
The statistic is summarised in Table \ref{tab:valid}.
The game instances recommended by RMHC and MABRMHC after optimising for OLMCTS with more iterations are still beneficial for the OLMCTS with less iterations. The game is still very difficult for the random agent. 

\begin{table}[h]
\centering
\caption{\label{tab:valid}Average winning rate (\%) over 11 recommendations after optimisation using 1,000 game evaluations, with different resampling numbers. Each game has been repeated 100 times.
}
\begin{tabular}{c|c|c}
\hline
Algorithm & 5 samples & 50 samples\\
\hline
RMHC & 86.00 & 91.8182\\
\hline
MABRMHC & 81.23 & 80.9545\\
\hline
\end{tabular}
\end{table}

\subsection{But what are the evolved games actually like?}

 To understand the results of the optimisation process, we visually inspected a random sample of games that had been found to have high fitness in the optimisation process, and compared these with several games that had low fitness.

We can discern some patterns in the high-fitness games. One of them is to simply have a very high cost for firing missiles. This is somewhat disappointing, as it means that the OLMCTS agent will score higher simply by staying far away from the RAS agent. The latter will quickly reach large negative scores.

A more interesting pattern was to have low missile costs, slow missiles, fast turning speed and fast thrusters. This resulted in a behaviour where the OLMCTS agent coasts around the screen in a mostly straight line, most of the time out of reach of the RAS agent's missiles. When it gets close to the RAS agent, the OLMCTS turns to intercept and salvo of missiles (which typically all hit), and then flies past.

In contrast, several of the low-fitness games have low missile costs and low cool-down times, so that the RAS agent effectively surrounds itself with a wall of missiles. The OLMCTS agent will occasionally attack, but typically loses more score from getting hit and than it gains from hitting the RAS agent. An example of this can be seen in Fig.~\ref{fig:ms}.

It appears from this that the high-fitness games, at least those that do not have excessive missile costs, are indeed deeper games in that skilful play is possible.

\section{Conclusion and further work}\label{sec:conc}

The work described in this paper makes several contributions in different directions.
Our main aim in this work is to provide an automatic game tuning method using simple but efficient black-box noisy optimisation algorithms, which can serve as a base-level game generator and part on an AI-assisted game design tool, assisting a human game designer with tuning the game for depth. The baseline game generator can also help with suggesting game variants that a human designer can build on.
Conversely, instead of initialising the optimisation with randomly generated parameters in the search space (as what we have done in this paper), human game designers can provide a set of possibly good initial parameters with their knowledge and experiences. 

The game instance evolving provides a method for automatic game parameter tuning or for automatically designing new games or levels by defining different fitness function used by optimisation algorithms. Even a simple algorithm such as RMHC may be used to automate game tuning. The application of other optimisation algorithms is straightforward.

The two tested optimisation algorithms achieve fast convergence towards the optimum even with a small resampling number when optimising for the OLMCTS against the RAS. 
Using dynamic non-adaptive or adaptive resampling numbers increasing with the generation number, such as the resampling rules discussed in \cite{liu2015portfolio}, to take the strength of both small and big numbers of resamplings will be favourable.

Though the primary application of MABRMHC to the space-battle game shows its strength, there is still more to explore. 
For instance, the selection between parent and offspring is still achieved by resampling each of them several times and comparing their average noise fitness value. 
However, the classic bandit algorithm stores the average reward and the times that each sampled candidate has been re-evaluated, which is also a form of resampling. We are not making use of this information while making the choice between the parent and offspring at each generation. Using a better recommendation policy (such as UCB or most visited) seems like a fruitful avenue of future work.
Another potential issue is the dependencies between the parameters to be optimised in some games or other real world problems. A N-Tuple Bandit Evolutionary Algorithm~\cite{kunanusont2017ntuple} is proposed to handle such case.

The study of the winning rate distribution and landscape over game instances helps us understand more about the game difficulty. Another possible future work is the study of fitness distance correlation across parameters. Isaksen et al.~\cite{isaksen2015discovering} used Euclidean distance for measuring distance between game instances of \emph{Flappy Bird} and discovered that such a simple measure can be misleading, since the difference between game instances does not always reflect the difference between their parameter values. We observe the same situation when analysing the landscape of fitness value by the possible values of individual game parameter (Fig. \ref{fig:ms}).

Though we focus on a discrete domain in this work, it's obviously applicable to optimise game parameters in continuous domains, either by applying continuous black-box noisy optimisation algorithms or by discretising the continuous parameter space to discrete values.
Evolving parameters for some other games, such as the games in GVG-AI framework, is another interesting extension of this work.

The approach is currently constrained by the limited intelligence of the GVG-AI agent we used, the proof of which is that on many instances of the game a reasonable human player is able to defeat both rotate-and-shoot (RAS) and the OLMCTS players. This problem will be overcome over time as the set of available GVG-AI agents grows.

\balance

\bibliographystyle{IEEEtran}

\bibliography{main}

\begin{thebibliography}{10}
\providecommand{\url}[1]{#1}
\csname url@samestyle\endcsname
\providecommand{\newblock}{\relax}
\providecommand{\bibinfo}[2]{#2}
\providecommand{\BIBentrySTDinterwordspacing}{\spaceskip=0pt\relax}
\providecommand{\BIBentryALTinterwordstretchfactor}{4}
\providecommand{\BIBentryALTinterwordspacing}{\spaceskip=\fontdimen2\font plus
\BIBentryALTinterwordstretchfactor\fontdimen3\font minus
  \fontdimen4\font\relax}
\providecommand{\BIBforeignlanguage}[2]{{%
\expandafter\ifx\csname l@#1\endcsname\relax
\typeout{** WARNING: IEEEtran.bst: No hyphenation pattern has been}%
\typeout{** loaded for the language `#1'. Using the pattern for}%
\typeout{** the default language instead.}%
\else
\language=\csname l@#1\endcsname
\fi
#2}}
\providecommand{\BIBdecl}{\relax}
\BIBdecl

\bibitem{togelius2008experiment}
J.~Togelius and J.~Schmidhuber, ``An experiment in automatic game design.'' in
  \emph{Proceedings of the 2008 IEEE Conference on Computational Intelligence
  and Games}, 2008, pp. 111--118.

\bibitem{browne2010evolutionary}
C.~Browne and F.~Maire, ``Evolutionary game design,'' \emph{IEEE Transactions
  on Computational Intelligence and AI in Games}, vol.~2, no.~1, pp. 1--16,
  2010.

\bibitem{cook2012aesthetic}
M.~Cook, S.~Colton, and A.~Pease, ``Aesthetic considerations for automated
  platformer design.'' in \emph{The Eighth Annual AAAI Conference on Artificial
  Intelligence and Interactive Digital Entertainment (AIIDE)}, 2012.

\bibitem{shaker2016procedural}
N.~Shaker, J.~Togelius, and M.~J. Nelson, ``Procedural content generation in
  games: A textbook and an overview of current research,'' \emph{Procedural
  Content Generation in Games: A Textbook and an Overview of Current Research},
  2016.

\bibitem{isaksen2015discovering}
A.~Isaksen, D.~Gopstein, J.~Togelius, and A.~Nealen, ``Discovering unique game
  variants,'' in \emph{Computational Creativity and Games Workshop at the 2015
  International Conference on Computational Creativity}, 2015.

\bibitem{liu2016rolling}
J.~Liu, D.~P{\'e}rez-Li{\'e}bana, and S.~M. Lucas, ``Rolling horizon
  coevolutionary planning for two-player video games,'' in \emph{Proceedings of
  the IEEE Computer Science and Electronic Engineering Conference (CEEC)},
  2016.

\bibitem{pell1992metagame}
B.~Pell, ``Metagame in symmetric chess-like games,'' 1992.

\bibitem{togelius2007towards}
J.~Togelius, R.~De~Nardi, and S.~M. Lucas, ``Towards automatic personalised
  content creation for racing games,'' in \emph{2007 IEEE Symposium on
  Computational Intelligence and Games}.\hskip 1em plus 0.5em minus 0.4em\relax
  IEEE, 2007, pp. 252--259.

\bibitem{togelius2011search}
J.~Togelius, G.~N. Yannakakis, K.~O. Stanley, and C.~Browne, ``Search-based
  procedural content generation: A taxonomy and survey,'' \emph{IEEE
  Transactions on Computational Intelligence and AI in Games}, vol.~3, no.~3,
  pp. 172--186, 2011.

\bibitem{hastings2009automatic}
E.~J. Hastings, R.~K. Guha, and K.~O. Stanley, ``Automatic content generation
  in the galactic arms race video game,'' \emph{IEEE Transactions on
  Computational Intelligence and AI in Games}, vol.~1, no.~4, pp. 245--263,
  2009.

\bibitem{sorenson2010towards}
N.~Sorenson and P.~Pasquier, ``Towards a generic framework for automated video
  game level creation,'' in \emph{European Conference on the Applications of
  Evolutionary Computation}.\hskip 1em plus 0.5em minus 0.4em\relax Springer,
  2010, pp. 131--140.

\bibitem{ashlock2010automatic}
D.~Ashlock, ``Automatic generation of game elements via evolution,'' in
  \emph{Proceedings of the 2010 IEEE Conference on Computational Intelligence
  and Games}.\hskip 1em plus 0.5em minus 0.4em\relax IEEE, 2010, pp. 289--296.

\bibitem{cook2011multi}
M.~Cook and S.~Colton, ``Multi-faceted evolution of simple arcade games.'' in
  \emph{Proceedings of the 2011 IEEE Conference on Computational Intelligence
  and Games}, 2011, pp. 289--296.

\bibitem{nelson2007towards}
M.~J. Nelson and M.~Mateas, ``Towards automated game design,'' in
  \emph{Congress of the Italian Association for Artificial Intelligence}.\hskip
  1em plus 0.5em minus 0.4em\relax Springer, 2007, pp. 626--637.

\bibitem{powley2016automated}
E.~J. Powley, S.~Gaudl, S.~Colton, M.~J. Nelson, R.~Saunders, and M.~Cook,
  ``Automated tweaking of levels for casual creation of mobile games,'' 2016.

\bibitem{lantz2017depth}
F.~Lantz, A.~Isaksen, A.~Jaffe, A.~Nealen, and J.~Togelius, ``Depth in
  strategic games,'' in \emph{under review}, 2017.

\bibitem{nielsen2015general}
T.~S. Nielsen, G.~A. Barros, J.~Togelius, and M.~J. Nelson, ``General video
  game evaluation using relative algorithm performance profiles,'' in
  \emph{European Conference on the Applications of Evolutionary
  Computation}.\hskip 1em plus 0.5em minus 0.4em\relax Springer, 2015, pp.
  369--380.

\bibitem{lucas2003learning}
S.~M. Lucas and T.~J. Reynolds, ``{Learning DFA: Evolution versus Evidence
  Driven State Merging},'' in \emph{Evolutionary Computation, 2003. CEC'03. The
  2003 Congress on}, vol.~1.\hskip 1em plus 0.5em minus 0.4em\relax IEEE, 2003,
  pp. 351--358.

\bibitem{lucas2005learning}
------, ``Learning deterministic finite automata with a smart state labeling
  evolutionary algorithm,'' \emph{Pattern Analysis and Machine Intelligence,
  IEEE Transactions on}, vol.~27, no.~7, pp. 1063--1074, 2005.

\bibitem{liu2016bandit}
J.~Liu, D.~Pe{\'r}ez-Liebana, and S.~M. Lucas, ``Bandit-based random mutation
  hill-climbing,'' \emph{arXiv preprint arXiv:1606.06041}, 2016.

\bibitem{liu2017banditrmhc}
D.~P.-L. Jialin~Liu and S.~M. Lucas, ``Bandit-based random mutation
  hill-climbing,'' in \emph{Evolutionary Computation, 2017. CEC'17. The 2017
  Congress on}.\hskip 1em plus 0.5em minus 0.4em\relax IEEE, 2017.

\bibitem{liu2016optimal}
J.~Liu, M.~Fairbank, D.~P{\'e}rez-Li{\'e}bana, and S.~M. Lucas, ``Optimal
  resampling for the noisy onemax problem,'' \emph{arXiv preprint
  arXiv:1607.06641}, 2016.

\bibitem{liu2015portfolio}
J.~Liu, ``Portfolio methods in uncertain contexts,'' Ph.D. dissertation,
  Universit{\'e} Paris-Saclay, 2015.

\bibitem{kunanusont2017ntuple}
K.~Kunanusont, R.~D. Gaina, J.~Liu, D.~Perez-Liebana, and S.~M. Lucas, ``The
  n-tuple bandit evolutionary algorithm for automatic game improvement,'' in
  \emph{Evolutionary Computation, 2017. CEC'17. The 2017 Congress on}.\hskip
  1em plus 0.5em minus 0.4em\relax IEEE, 2017.

\end{thebibliography}
\end{document}